# The Nature of the unnormalized Beliefs encountered in the Transferable Belief Model.

**Philippe Smets**
IRIDIA
Université Libre de Bruxelles
Brussels-Belgium

**Abstract:**

Within the transferable belief model, positive basic belief masses can be allocated to the empty set, leading to unnormalized belief functions. The nature of these unnormalized beliefs is analyzed.

## 1. INTRODUCTION.

In subjective probability theory, a probability space $(\Omega, \mathcal{A}, P)$ is characterized by a set of worlds $\Omega$, a Boolean algebra $\mathcal{A}$ of subsets of $\Omega$ and a probability measure P defined on $\mathcal{A}$. Let $\varpi$ be the world that corresponds to the actual sate of affairs. We ignore which world corresponds to $\varpi$. We only know the strength P(A) of our belief that $\varpi \in A$, for all $A \in \mathcal{A}$. The normalization of P is an axiom of the theory: $P(\Omega) = 1$.

In the transferable belief model (TBM), the model we developed to quantify beliefs and that covers the same domain of application as the subjective probability measures, we distinguish between an open-world and a close-world context (Smets 1988, Smets and Kennes 1990). In the open-world context beliefs are not necessarily normalized. The nature of this possible lack of normalization is analysed in this paper.

The TBM is characterized by an unitary mass of belief that is distributed among the subsets of a finite frame of discernment $\Omega$. For $A \subseteq \Omega$, m(A) is the portion of our total initial belief supporting $A^1$ and that could support any subset of A if further information justifies it, but given the available information m(A) supports only A. The function $m:2^{\Omega} \rightarrow [0,1]$ is called a basic belief assignment (bba) and the m values are called the basic belief masses (bbm), with

$$\sum_{A \subseteq \Omega} m(A) = 1$$

Based on the bba m, the functions bel(A) and pl(A) are defined for $A \subseteq \Omega$ by:

$$bel(A) = \sum_{\emptyset \neq X \subseteq A} m(X) \tag{1}$$

$$pl(A) = \sum_{A \cap X \neq \emptyset} m(X) \tag{2}$$

$$bel(\emptyset) = pl(\emptyset) = 0.$$

Hence $\quad pl(A) = bel(\Omega) - bel(\overline{A})$

$$bel(\Omega) = pl(\Omega) = 1 - m(\emptyset) \leq 1.$$

For all $A \subseteq \Omega$, bel(A) (read the belief of A) represents the degree of *justified specific support* given to A and pl (A) (read the plausibility of A) represents the degree of *potential specific support* that could be given to A.

We say *specific* because the bbm's m(X) included in bel(A) are those that support A without supporting $\overline{A}$. Hence X must be a subset of A without being a subset of $\overline{A}$, i.e. $X \neq \emptyset$. So m($\emptyset$) is not included in bel(A). Identically it is not included in pl(A) as pl(A) includes the bbm given to sets X compatible with A, and $\emptyset$ is not compatible with A.

---

[1] To support A means to support $\varpi \in A$.

We say *justified* because *only* the bbm's given to subsets of A are included in bel(A). For instance, consider two distinct elements x and y of $\Omega$. The bbm $m(\{x,y\})$ given to $\{x,y\}$ could support x if further information indicates this. However given the available information the bbm can only be given to $\{x,y\}$.

We say *potential* because the bbm included in pl(A) could be transferred to non empty subsets of A if some new information could justify such a transfer. It would be the case if we learn that all the worlds in $\overline{A}$ are impossible. The fact that, after such a conditioning, the updated belief given to A will be equal to pl(A) reflects the fact that pl(A) is the maximal support that might be given to A. In particular, subsets $X \subseteq \Omega$ for which it is accepted that $\varpi \notin X$ receive a null plausibility. If it was not the case, then appropriate updating could induce a positive support to X, thus making supported worlds that were accepted as not including $\varpi$.

An important element of the TBM is the acknowledgement that a positive bbm can be allocated to the empty set or equivalently that bel($\Omega$) and pl($\Omega$) can both be less than one. The bbm $m(\varnothing)$ can be viewed as a *missing mass* as it is equal to $1$-pl($\Omega$). On the contrary, Shafer (1976) defines a belief function such that $m(\varnothing)=0$ or equivalently bel($\Omega$)=pl($\Omega$)=1.

In order to analyse the nature of $m(\varnothing)>0$, we define the nature of the frame of discernment in section 2. In section 3, we explain the conditioning process used in the TBM. In section 4, we explain the origin of the positive bbm given to $\varnothing$. In section 5, we analyse the case where the frame of discernment is not exhaustive. We summarize the results in section 6.

## 2. THE FRAME OF DISCERNMENT.

In the TBM, bel quantifies the strength of the beliefs held by a given agent Y at a given time t that an actual state of affairs $\varpi$ belongs to subsets of possible worlds. The domain of Y's beliefs at time t is a set of distinct possible worlds, one of them, denoted $\varpi$, corresponds to the actual state of affairs (Carnap 1956, 1962, Ruspini 1986). This set, is denoted by $\Omega_L$ and called the frame of discernment. It is defined as follows.

## 2.1. THE LOGICAL CONSTRUCTION OF $\Omega_L$.

Let L be a finite propositional language. Let $\Omega_L = \{\omega_1, \omega_2, ...\omega_n\}$ be the set of worlds that correspond to the interpretations of L. Propositions identify the subsets of $\Omega_L$. Let $\top$ be the tautology and $\bot$ be the contradiction. For any proposition X, let $[\![X]\!] \subseteq \Omega_L$ be the set of worlds identified by X. Let A be a subset of $\Omega_L$, then $f_A$ is any proposition that identifies A. So $A=[\![f_A]\!]$, $\varnothing=[\![\bot]\!]$ and $\Omega_L=[\![\top]\!]$. The domain of bel and pl are sets of worlds in $\Omega_L$. By definition the actual world $\varpi$ is an element of $\Omega_L$.

Given a frame of discernment $\Omega_L$, the complement $\overline{A}$ of a set of worlds $A \subseteq \Omega_L$ is the complement of A relative to $\Omega_L$. It corresponds to $[\![\neg f_A]\!]$.

## 2.2. THE EVIDENTIAL CORPUS.

The beliefs held by Y on $\Omega_L$ at time t are based on the set of all the pieces of evidence held by Y at time t. This set os called the evidential corpus and denoted $EC_t^Y$. Updating of $EC_t^Y$ can be obtained by adding a new piece of evidence to $EC_t^Y$. Let t'>t and let Ev be the piece of evidence added to $EC_t^Y$. Then $EC_{t'}^Y = EC_t^Y \cup \{Ev\}$. Ev induces a conditioning of Y's beliefs on $\Omega_L$ if that piece of evidence is relevant to Y's beliefs on $\Omega_L$. The evidential corpus $EC_t^Y$ can also be updated by the removing of a piece of evidence from $EC_t^Y$. Then Ev$\in EC_t^Y$ and Ev$\notin EC_{t'}^Y$. It results in a deconditionalization of Y's beliefs on $\Omega_L$ if that piece of evidence was relevant to Y's beliefs on $\Omega_L$ at time t (Klawonn and Smets 1992). These two kinds of changes correspond to the expansion and contraction processes (Levi , 1980, pg. 25)

## 3. CONDITIONING IN THE TBM.

In this section, we describe the impact of a new pieces of evidence within the TBM. Let $\Omega_L$ be a frame of discernment. A piece of evidence $I_A$ is added to $EC_t^Y$ and $I_A$ is such that Y takes for grant that $\varpi \in A$.

In the TBM, the bbm given to $X \subseteq \Omega_L$ is transferred to $X \cap A$. When $X \cap A \neq \varnothing$, this transfer reflects the nature of the TBM and its masses: a bbm given to a set X is a part of belief that supports X and might support any subset of X if further information justifies such a support to a more specific subset. The information that the worlds in $\overline{A}$ are impossible is the kind of information that justifies such a transfer. Y updates his beliefs by conditioning m ( or equivalently bel and pl) on A. The resulting bba $m_A$ is:

$$m_A(B) = \begin{cases} \sum_{X \subseteq \overline{A}} m(B \cup X) & \forall B \subseteq A, \\ 0 & \text{otherwise} \end{cases}$$

Thus    $m_A(\emptyset) = m(\emptyset) + bel(\overline{A})$.

$bel_A$, $pl_A$ are related to $m_A$ by relations (1) and (2). The results are equivalent to:

$$bel_A(B) = bel(B \cup \overline{A}) - bel(\overline{A}) \qquad \forall B \subseteq \Omega_L$$

and    $pl_A(B) = pl(A \cap B) \qquad \forall B \subseteq \Omega_L$.

This updating corresponds to the unnormalized Dempster's rule of conditioning. Justifications of this rule can be found in Nguyen and Smets (1991) and in Klawonn and Smets (1992).

## 4. UPDATING BELIEFS.

Let $\Omega_L$ be the frame of discernment on which Y builds his/her beliefs at time t. Let m be the bba that quantifies these beliefs that result from $EC_t^Y$. We want to explain the meaning of $m(\emptyset)>0$ as encountered in the TBM updating.

**4.1.** Let us suppose that Y learns the piece of evidence $I_A$ that the worlds in $\overline{A}$ happen to be impossible, with $A \subseteq \Omega_L$, $A \neq \emptyset$. Let bel' be the result of the updating of bel by this pieces of evidence, whereas $bel_A$ is the updating obtained in the TBM. bel' is the belief function that results from $EC_t^Y \cup \{I_A\}$. We give our reason why bel' should be $bel_A$.

The transfer of the bbm $m(X)$, $X \subseteq \Theta$ to $X \cap A$ is accepted when $X \cap A \neq \emptyset$, as this transfer is at the core of the TBM. What has to be justified is the transfer of $bel(\overline{A})$ to $m_A(\emptyset)$ and the non-normalization.

$bel(\overline{A})$ is the sum of the bbm given to the non-empty subset of $\overline{A}$. The updating information $I_A$ says that the worlds in $\overline{A}$ are impossible. Therefore $bel(\overline{A})$ was a support given by Y at time t to a set of worlds that turns out to be impossible. It results from a conflict between the beliefs on $\Omega_L$ induced by the $EC_t^Y$ and responsible for Y's allocation of positive bbm to some non-empty subsets of $\overline{A}$ and the beliefs on $\Omega_L$ induced by the new piece of evidence $I_A$ that says that no positive bbm should be allocated to the non-empty subsets of $\overline{A}$. So $bel(\overline{A})$ quantifies the conflict between Y's beliefs at time t and the beliefs induced by the updating piece of evidence. It must be eliminated, either by a transfer to some kind of 'absorbing world' that represents a 'contradictory' state, denoted as $\phi$, or by being redistributed among the still possible worlds by some normalization process.

Two types of arguments can be used to show that the redistribution of $bel(\overline{A})$ among the worlds included in A is inappropriate. The first is based on the definition of the plausibility, the second on a homomorphisme requirement.

For any subset X of the frame of discernment, we defined $pl(X)$ as the degree of potential specific support that could be given to X. By potential we mean that $pl(X)$ represents the maximal degree of justified support that might be allocated to X. Being maximal, $pl(X)$ can not increase after conditioning. This explains why we require that $pl_A(A) = pl(A)$, in which case the redistribution of $bel(\overline{A})$ among the subsets of A is not allowed. This argument can be questioned as it is based on a particular definition of the plausibility .

For the homomorphisme requirement (Gärdenfors, 1988), consider two belief functions bel' and bel" defined on a frame of discernment $\Omega_L$ and a random device which outcome indicates which belief function is selected. Let p be the probability of bel' being selected and let q=1-p be the probability that bel" is selected. Let bel be the belief function on $\Omega_L$ resulting from the overall schema, so $bel(X) = p \ bel'(X) + q \ bel"(X) \ \forall X \subseteq \Omega_L$ (proved in Smets 1990b). The homomorphisme requirement states that the same relation should hold whatever the conditioning subset:

**Homomorphisme requirement :**

If    $bel(X) = p \ bel'(X) + (1-p) \ bel"(X)$ ,
      $\forall X \subseteq \Omega_L$, $p \in [0,1]$,

then    $bel_A(X) = p \ bel'_A(X) + (1-p) \ bel"_A(X)$
      $\forall X \subseteq \Omega_L$, $\forall A \subseteq \Omega_L$

The homomorphisme is satisfied iff $bel(\overline{A})$ is not redistributed among the subsets of A.`

**Proof**: In general $bel_A(X) = bel(X \cup \overline{A}) - bel(\overline{A}) + g(bel(\overline{A}))$. The first and second terms corresponds to the transfer of the bbm given to the subsets of $\Omega_L$ compatible with A. The third is the fraction of $bel(\overline{A})$ due to the hypothetical redistribution of $bel(\overline{A})$ among the subsets of A. The homomorphisme requirement becomes: for $p \in [0,1]$,

$$\forall X \subseteq \Omega_L, \quad bel_A(X) = p \ bel'_A(X) + (1-p) \ bel''_A(X)$$

or

$$bel(X \cup \overline{A}) - bel(\overline{A}) + g(bel(\overline{A})) =$$
$$p \ [bel'(X \cup \overline{A}) - bel'(\overline{A}) + g'(bel'(\overline{A}))] +$$
$$(1-p) \ [bel''(X \cup \overline{A}) - bel''(\overline{A}) + g''(bel''(\overline{A}))]$$

where the g functions can depend on the belief function considered.

Given $bel(X) = p \ bel'(X) + (1-p) \ bel''(X)$, the last relation becomes:

$$g(p \ bel'(\overline{A}) + (1-p) \ bel''(\overline{A})) =$$
$$p \ g'(bel'(\overline{A})) + (1-p) \ g''(bel''(\overline{A}))$$

or $\quad g(px + qy) = p \ g'(x) + q \ g''(y)$

which solutions for g, g' and g" are linear functions (Aczel, 1966, pg. 144).

So $\quad g(x) = a \ x + b$, with a and b constant,

and $\quad bel_A(X) = bel(X \cup \overline{A}) - bel(\overline{A}) + a \ bel(\overline{A}) + b$.

The requirement that $bel_A(\overline{A}) = 0$ implies that: a $bel(\overline{A})$ + b = 0 whatever $bel(\overline{A})$, hence a = b = 0.     QED

The satisfaction of the homomorphisme requirement or the definition of the plausibility function imply that $bel(\overline{A})$ can not be redistributed among the subsets of A. So $bel(\overline{A})$ can only be transferred to some contradictory state φ or to Ø.

We show now that there is at most one contradictory state φ. We accept that the updating process is the same when applied to a belief function defined on $\Omega_L$ or on the coarsening of $\Omega_L$ obtained by keeping the elements in A and regrouping the elements in $\overline{A}$ into a single new element. We accept that the updated beliefs obtained in both frames are the same. It means that the 'contradictory' states absorbing the bbm given to the subsets of $\overline{A}$ in the initial frame of discernment is identical to the 'contradictory' state absorbing the bbm given to the single element representing $\overline{A}$ in the coarsening of the initial frame of discernment. This is true whatever $A \subseteq \Omega_L$. Therefore there is an unique contradictory state φ.

We show finally that φ=Ø. By definition, bel(X) and pl(X), $X \subseteq \Omega_L$, include only those bbm given to non-empty subsets of X or non-empty subsets compatible with X, respectively. Even if φ≠Ø, φ may not be a subset of any subset of A, otherwise the bbm $bel(\overline{A})$ that was transferred to φ would be added into the beliefs given to these subsets of A, contrary to the fact that $bel(\overline{A})$ may not be redistributed among the subsets of A. So φ⊄A. It may neither be a subset of $\Omega_L$ compatible with $\overline{A}$ as otherwise $pl_A(\overline{A})$ would become positive in contradiction to the fact that all worlds in $\overline{A}$ are assumed to be impossible. So φ⊄$\Omega_L$, what is impossible if φ≠Ø. Therefore the only remaining solution is φ=Ø, in which bel' = $bel_A$, and $bel(\overline{A})$ is thus transferred to $m_A(Ø)$.

**4.2.** We still have to consider the case A=Ø, i.e.the case where all the worlds in $\Omega_L$ are impossible. Y would be in a state of complete contradiction. It is a situation analogous to the one encountered in probability theory when conditioning on an event of zero probability. No natural solution exists and therefore could be imposed, hence there are no criteria to test the solution. In the TBM, such state of complete contradiction is translated by m(Ø)=1, a bba that naturally translates the state of complete contradiction to be characterized.

We have argued that conditioning on the information $I_A$ = 'the worlds in $\overline{A}$ are impossible' can result in the transfer of $bel(\overline{A})$ to $m_A(Ø)$. The information $I_A$ was added to $EC_t^Y$, Y's evidential corpus at time t, and $m_A$ was the resulting bba induced by $EC_t^Y \cup \{I_A\}$. Before $I_A$ was added to $EC_t^Y$, how to justify that the bbm m(Ø) induced by $EC_t^Y$ could already be positive? It is due to the fact that the bba induced by $EC_t^Y$ might already result from a conditioning on the pieces of evidence that were present in $EC_t^Y$. Contradiction could already be present in $EC_t^{Y\,2}$. Of course, contradiction within $EC_t^Y$ is not necessarily present, in which case $bel(\Omega_L)=1$. Once pieces of evidence accumulate in the evidential corpus, some contradiction will soon appear. Y can get rid of this contradiction by a deconditionalization process, the process by which the pieces of evidence in $EC_t^Y$ that were responsible for the contradiction are excluded from $EC_t^Y$. The concept of deconditionalization in defined in Klawonn and Smets (1992).

In conclusion, the bbm $m(\emptyset)$ quantifies the amount of conflict present in Y's beliefs induced by Y's evidential corpus $EC_t^Y$ at time t. Renormalization would remove this conflict. This solution permits to hide conflicts but might be misleading as as shown in the example described by Zadeh (1984).

## 5. THE EPISTEMIC CONSTRUCT OF THE FRAME OF DISCERNMENT.

We already studied the belief induced on a frame of discernment $\Omega_L$ derived from a propositional language L. In practice, the frame of discernment can also be built by listing the possible worlds, without referring explicitly to an underlying propositional language L. The frame of discernment is then an epistemic construct.

For this epistemic construct, one acknowledges the "limited understanding" of the agent Y that holds the beliefs (Walley, 1991). Y establishes a list $\Omega_t^Y$ of possible worlds, like the list encountered in a database. Because of Y's "limited understanding", Y is not able to imagine *all* the possible worlds. $\Omega_t^Y$ contains only those worlds that Y can conceive of at time t. The indices in $\Omega_t^Y$ emphasize the time dependency and the personal nature of the frame of discernment. Let $\Delta_t^Y$ denote the set of worlds not conceived by Y at t, but that could have been conceived if a logical approach had been taken, i.e. if one had started with the language L that underlies the worlds in $\Omega_t^Y$. Thus $\Omega_t^Y \cup \Delta_t^Y = \Omega_L$ where $\Omega_L$ is the set of worlds that correspond to the interpretations of L.

Y's beliefs can be expressed only for the worlds that he can conceived of, hence those in $\Omega_t^Y$. By definition, Y do not conceive the worlds in $\Delta_t^Y$, and therefore Y surely cannot express his beliefs that the actual state of affairs $\varpi$ corresponds to one of the worlds in $\Delta_t^Y$. Y's beliefs at time t are only expressed for the subsets of worlds in $\Omega_t^Y$.

By building $\Omega_t^Y$, Y acknowledges implicitly - and maybe erroneously - that $\varpi$ belongs to $\Omega_t^Y$, therefore implicitly conditioning on $\Omega_t^Y$ a belief function that could have been defined on $\Omega_L$. So all the bbm are allocated to subsets of $\Omega_t^Y$. As contradiction could already result from this conditioning, some $m(\emptyset)>0$ might be encountered. It correspponds to the belief that would have been allocated to $\Delta_t^Y$.

## 6. CONCLUSIONS.

In summary, we have shown the nature of $m(\emptyset)>0$. The bbm given to $\emptyset$ at time t quantifies the amount of contradiction present at time t in the belief function that quantifies Y's beliefs about the set of worlds that are conceivable for him at time t.

A merit of the TBM is that it keeps track of the contradiction present in the $EC_t^Y$. Renormalization as done in the Shafer's approach leads to unnatural results like the one described by Zadeh (1984). Counterexamples like Zadeh's explain why we gave up the normalization requirement when we start developing the TBM[3].

Two questions about the meaning of $m(\emptyset)>0$ merit consideration.

**6.1.** Suppose Y's initial beliefs are such that $bel(\varpi\in\Omega_L)$ $=1$, and Y learns that $\varpi\notin\overline{A}$ by adding the evidence $I_A$ to $EC_t^Y$ at t'>t: so $EC_{t'}^Y=EC_t^Y\cup\{I_A\}$. How come that $bel(\varpi\in A)$ is not necessarily 1? The answer is to be found in the definition of bel: bel is the degree of justified support, and $m(\emptyset)$ is the amount of inherent contradiction present among the beliefs induced on $\Omega$ by Y's evidential corpus $EC_t^Y$. bel does not quantify unwarranted beliefs. For instance, whims are not included in bel. bel quantifies warranted supports and the evidential corpus might be such that it supports $\varpi \in A$ somehow, but not fully, because it happens that $EC_t^Y$ and $I_A$ induce some contradictory supports on $\Omega$.

---

[2] This is not the only possible origin for a positive bbm given to $\emptyset$. It may also results from the application of Dempster's rule of combination, what is equivalent to a conditioning on a dubious piece of evidence.

[3] Historically this occurs during a discussion I had with L. Zadeh and R. Yager in a beautiful exotic restaurant in Acapulco in 1980.

The simple example can be encountered if $Ev_1 = 'X_1$ says $\varpi \in A'$ and $Ev_2 = 'X_2$ says $\varpi \in \overline{A}'$, where $X_1$ and $X_2$ are two different induviduals. Let $EC_t^Y = \{Ev_1\} \cup \{Ev_2\}$. $Ev_1$ justifies Y's beliefs that $\varpi \in A$, and $Ev_2$ justifies Y's beliefs that $\varpi \in \overline{A}$. Both beliefs/supports are contradictory, and $m(\emptyset)=1$ translates the fact that $EC_t^Y$ actually does not justify any support for A or $\overline{A}$.

Historically, Shafer (1976) already qualified the bbm we allocate to $\emptyset$ as the 'weight of conflict'. A high conflict would justify reconsidering how the evidential corpus influence beliefs (Laskey and Lehner, 1989). In the last example, where $m(\emptyset)=1$, it is obvious that Y should reconsider the impact of the two pieces of evidence $Ev_1$ and $Ev_2$, trying to discount or eliminate the 'most unreliable' one. When $1 > m(\emptyset) > 0$, the largest $m(\emptyset)$ the more seriously Y should reconsider how his evidential corpus influence his beliefs, but there is of course no real crisp limit between values of $m(\emptyset)$ that would correspond to acceptable conflict, and those that would not. It is a matter of degree. The way to update the impact of the evidential corpus in order to reduce the conflicts on $\Omega$ require information external to the model. It is not considered in this paper.

**6.2.** The second question is concerned with beliefs induced by randomness. Suppose a random device generates the mutually exclusive and exhaustive events $\omega_i$, i=1, 2, ...n with probabilities $p_i$. Should we quantify our belief that event $\omega_i$ will occur as $bel(\omega_i) = p_i$? Does randomness warrants support? These questions are left open. It seems that the $p_i$'s are the pignistic probabilities induced on $\Omega$ by an underlying belief on $\Omega$ (Smets 1990b). If the $p_i$'s are not equal, it seems to reflect that the evidential corpus justifies some beliefs on some subset of $\Omega$.

The aim of this paper was not to solve all the open questions in relation to $m(\emptyset) > 0$, but to provide a meaning for it. There are still pending questions that will be studied in future papers.

**Acknowledgments.** The author is indebted to M. Clarke, Y-T Hsia, R. Kennes, R. Kruse and V. Poznansky for their constructive remarks and their stimulating comments.